\documentclass[10pt,twocolumn,letterpaper]{article}

\usepackage[pagenumbers]{cvpr} %

\usepackage{graphicx}
\usepackage{amsmath}
\usepackage{amssymb}
\usepackage{booktabs}
\usepackage{multirow}
\usepackage{color, colortbl}
\usepackage{caption}

\usepackage[pagebackref,breaklinks,colorlinks]{hyperref}

\usepackage[capitalize]{cleveref}
\crefname{section}{Sec.}{Secs.}
\Crefname{section}{Section}{Sections}
\Crefname{table}{Table}{Tables}
\crefname{table}{Tab.}{Tabs.}

\def\blue#1{\textcolor{blue}{#1}}

\def\cvprPaperID{20} %
\def\confName{CVPR}
\def\confYear{2022}

\begin{document}

\title{Object Instance Identification in Dynamic Environments}

\author{Takuma Yagi, Md. Tasnimul Hasan, and Yoichi Sato\\
Industrial Institute of Science, The University of Tokyo\\
{\tt\small \{tyagi,tasnim,ysato\}@iis.u-tokyo.ac.jp}
}
\maketitle

\begin{abstract}
We study the problem of identifying object instances in a dynamic environment where people interact with the objects. In such an environment, objects' appearance changes dynamically by interaction with other entities, occlusion by hands, background change, etc. This leads to a larger intra-instance variation of appearance than in static environments. To discover the challenges in this setting, we newly built a benchmark of more than 1,500 instances built on the EPIC-KITCHENS dataset which includes natural activities and conducted an extensive analysis of it. Experimental results suggest that (i) robustness against instance-specific appearance change (ii) integration of low-level (\eg, color, texture) and high-level (\eg, object category) features (iii) foreground feature selection on overlapping objects are required for further improvement.
\end{abstract}

\section{Introduction}
\label{sec:intro}
The ability to identify whether an object appearing in an image is the same instance as previously seen is one of the essential abilities to understand physical environments over time.
For example, when a robot is instructed by the user to ``bring my phone'', the robot needs to identify {\it the phone} that has observed in the past from its appearance.
This object instance identification task offers several applications such as product search, robot control, and assistive technology.

\begin{figure}[t]
\centerline{\includegraphics[width=0.9\linewidth]{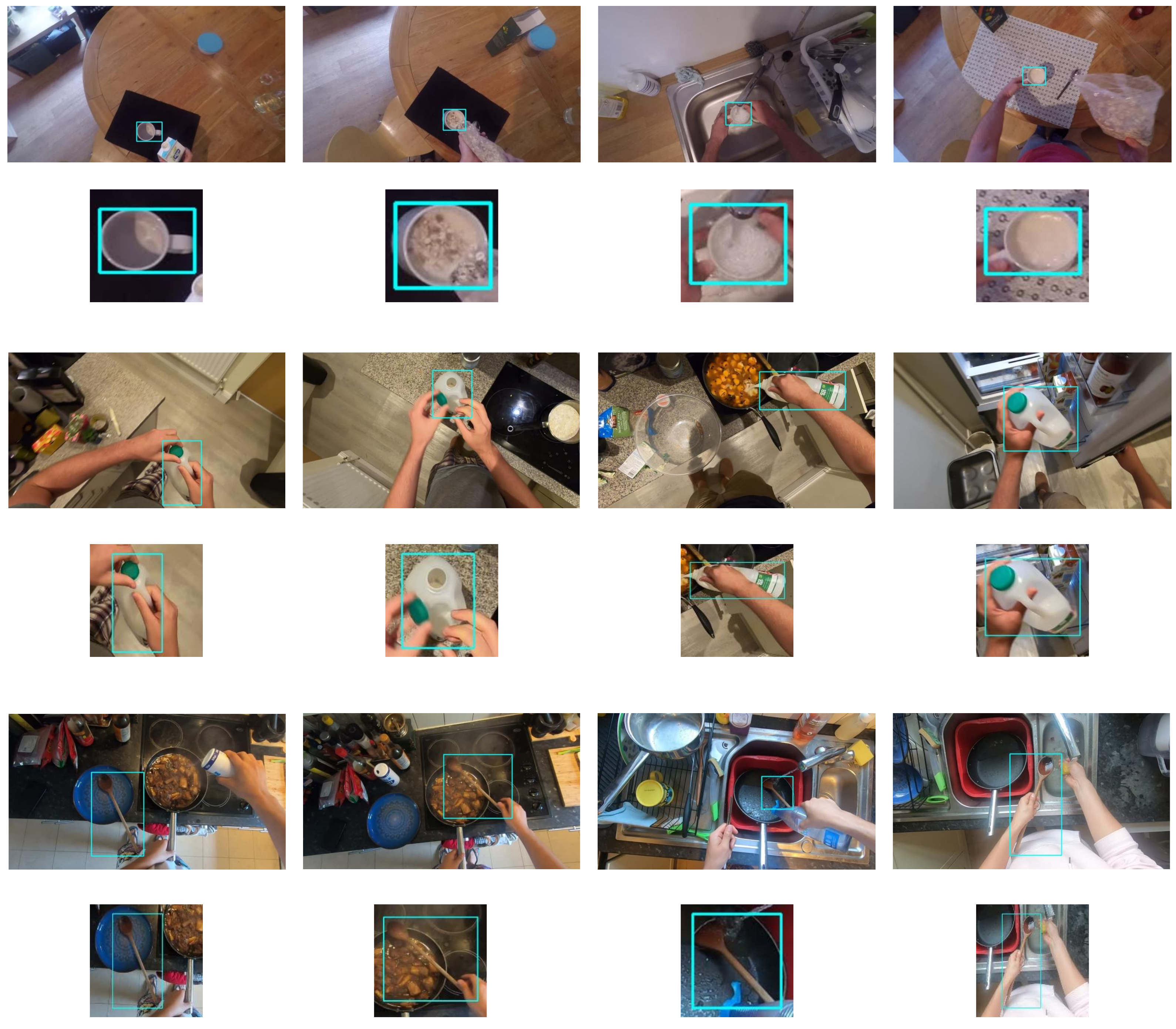}}
\caption{Examples of EK-Instance dataset. Each row denotes images from same instance.}
\label{fig:teaser_reid}
\end{figure}

Previously, recognition of object instances has been primarily studied in a static environment where the appearance of the object does not change across time.
However, in the real-world, objects are used for their intended use, and their appearance change through interactions.
For example, a mug may be filled with liquids or washed by hands during its use, causing significant foreground and background clutters (Figure~\ref{fig:teaser_reid}, top row).
Although identifying instances in such a {\it dynamic environment} is demanded, challenges in this situation has not been seriously discussed yet.

We analyze the problem of object instance identification under dynamic environment where people interact with objects.
To this end, we contribute a new annotation built on the large-scale EPIC-KITCHENS dataset~\cite{Damen2020RESCALING} that contains natural cooking activities.
We choose first-person videos as a subject because it contains realistic hand-tool use and habitual activities across times.
The new EK-Instance dataset (Figure~\ref{fig:teaser_reid}) consists of more than 1,500 object instances, and contains various clutter in the foreground (\eg, pouring drinks) and background (\eg, overlap with other objects).

To discover the challenges in this setting, we introduce baseline methods using the off-the-shelf metric learning and clustering, and conduct (1) performance evaluation in EK-Instance dataset (2) cross-dataset transfer for analysis.
The results suggest that (i) robustness against instance-specific appearance change (ii) integration of low-level and high-level features (iii) foreground feature selection are required for further improvement.

\section{Related Work} %
Instance recognition/detection in a monotonous background~\cite{held2016robust} and re-identification in a static environment~\cite{bansal2021where} have been studied, aiming for applications such as product recognition and robot control. 
Numbers of instance-level datasets have been proposed in parallel.
Their source is roughly divided into (i) recording under controlled environments and (ii) Internet images.
In the first group, at most, a few hundred instances were collected on a turntable or by hand-held.
Robustness against pose change and lighting conditions is their primary interests~\cite{wang2018toybox}.
Except for a few datasets that intentionally collected varied backgrounds~\cite{wang2015instre,lomonaco2017core50}, most datasets are captured in a controlled environment where objects are captured without interaction with monotonous background.
The second group collects data from E-Commerce sites for product recognition.
Compared to the former, it is easier to scale up the number of instances to tens of thousands of instances (\eg, \cite{song2016deep}).
However, for their original purpose, objects were shot with a simple background without real-world clutters.
Different from the above, we collect a large number of object instances that appear in natural environments with interaction.

\section{EK-Instance Dataset}
\begin{figure}[t]
\centerline{\includegraphics[width=1.0\linewidth]{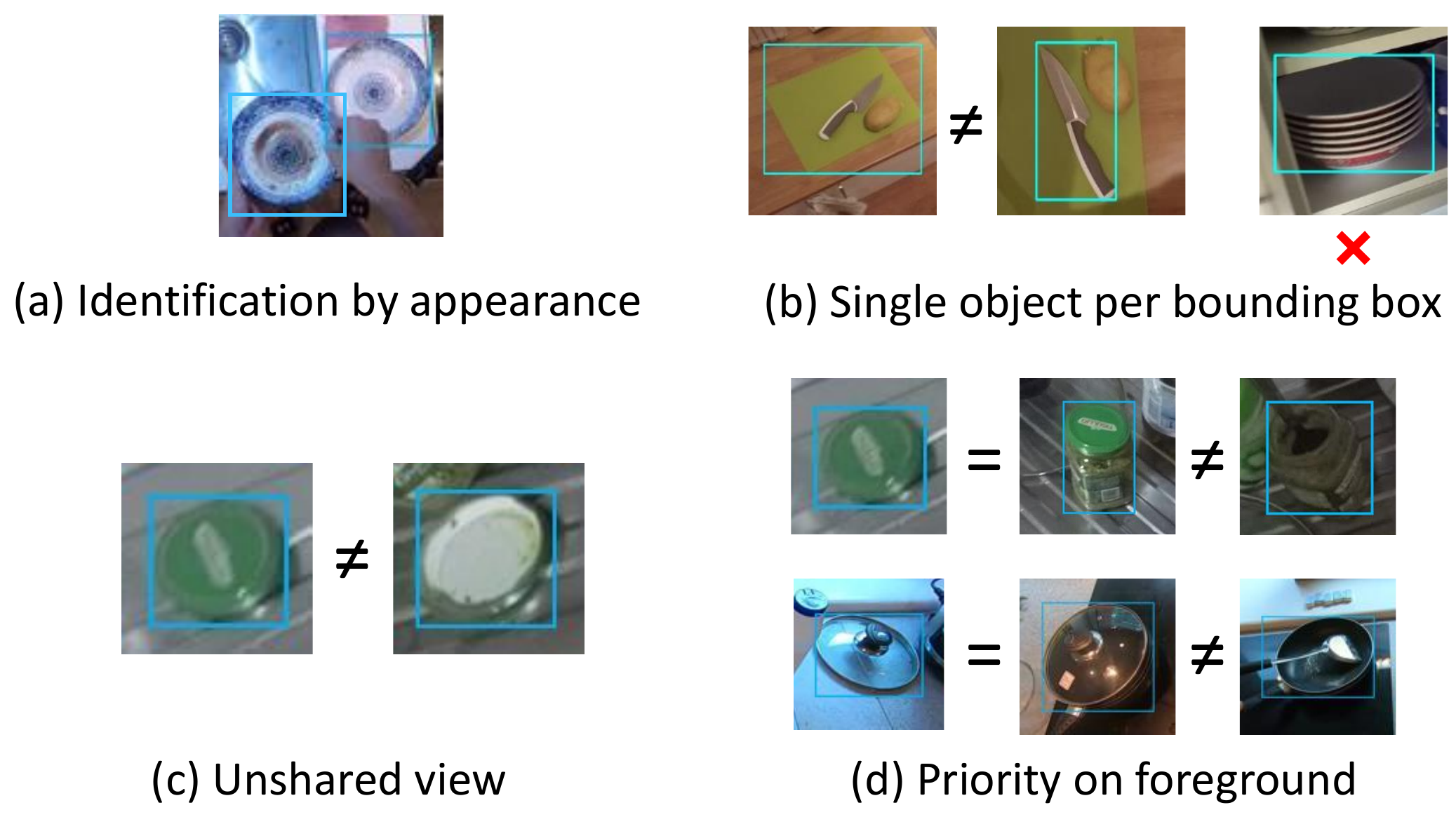}}
\caption{Criteria for determining an instance.}
\label{fig:principle}
\end{figure}

\subsection{Definition of Instance}
We first clarify what {\it an instance} refers to.
Compared to the previous works that captured objects apart from clutters, real-world environments include corner cases caused by interaction and it is not trivial how to determine the same instance.
We propose four rules for consistent annotation.

{\bf (a) Identification by appearance:}
We define an instance as {\it an object which has a unique appearance that is distinguishable from other objects}.
Visually identical objects (\eg, set of tableware) are counted as a single instance.

{\bf (b) Single object per bounding box:}
We use a bounding box that surrounds the object just enough on determining the target object.
When part of an object is occluded, we just cover the visible region.
If a bounding box surrounds multiple objects (\eg, stacked dishes), it is considered invalid and excluded from the evaluation (see Figure~\ref{fig:principle} (b)).

{\bf (c) Exception on unshared view:}
Ideally, a model should identify an object as the same even if it is viewed from different directions.
However, as shown in Figure~\ref{fig:principle} (c), in some cases, it will be difficult to do so when there is no common part across views, such as the colors of the front and back of the lid being completely different.
We exceptionally count as different instances in such a situation.

{\bf (d) Priority on foreground:}
Some objects can be separated into multiple parts.
For example, a frying pan can be separated into a body and a lid.
If we look at it from a top-down view, it will be impossible to distinguish between (i) a set of body and lid and (ii) the lid itself.
While multiple interpretations are possible, we give priority to the part on the foreground and count as the same if objects share the foreground.
In Figure~\ref{fig:principle} (d), we consider (i) and (ii) as the same while counting the body alone as a different instance.

\subsection{Annotation Procedure}
Based on the aforementioned criteria, annotations were made on the EPIC-KITCHENS dataset~\cite{Damen2020RESCALING} which includes continuous cooking activities across dates.
Because it is too labor-intensive to manually annotate instances across long videos, we adopt a semi-automatic procedure for annotation.
Specifically, a category-agnostic object detector~\cite{kim2021learning} and an object tracker are used to automatically extract and track the objects.
Then we manually inspect whether the obtained trajectories contain a valid object and the same instance as other tracks.
Finally, we sample detections by 1~fps and corrected the bounding boxes.

As a result, we collected 1,554 object instances, 38,479 tracks, and 92,376 frames from the 20 hours of video from 23 camera wearers in the EPIC-KITCHENS dataset.
It consisted of containers (\eg, bottle), cooking tools (\eg, frying pan and paper roll), tableware (\eg, knife), food (\eg, apple), and electronics (\eg, toaster).
It also included objects which are difficult to define in a clear category (\eg, fridge magnet, portafilter, and product package).

\section{Experiments}
Using the EK-Instance dataset, we evaluated the instance identification task in dynamic environments by building baseline models and comparing them across datasets.

\subsection{Problem Statement}
The problem is formulated as clustering of image tracks to a group of instances.
Given a set of image tracks $\{X_1, \dots, X_N \}$, we assign cluster labels $ \{y_1, \dots, y_N \} (y_i \in \{1, \dots, K \})$, which is grouped by instance.
In a novel environment, the true number of instances will not given in advance.
Therefore, we formulated the task as clustering.

\subsection{Image Track Encoder}
We use ResNet-34~\cite{he2016deep} pre-trained with the ImageNet dataset~\cite{deng2009imagenet} as a backbone network to extract frame-level features.
We pass the images to the backbone layers before the final average pooling layer and obtain a 512-dimensional feature vector for each image.
Next, they are averaged among images to aggregate image-level features into a track-level feature vector.
Finally, we pass it to a single fully-connected layer to obtain a 256-dimensional track embedding.
We used two loss functions for comparison:

{\bf Additive angular margin (ArcFace)~\cite{deng2019arcface}:} Loss based on classification. Dicriminative features are learned by giving a penalty that the feature vectors between different classes are separated from each other by a certain amount or more.

{\bf N-pair~\cite{sohn2016improved}:} Loss based on sample-wise comparison. Increase the similarity of pairs of feature vectors in the same class while decreasing the similarity of different classes.

In addition, we evaluated the {\bf ImageNet} baseline which extracts 512-dimensional feature vectors from ImageNet pre-trained model as a strong baseline model.

\subsection{Clustering and Evaluation}
We performed clustering using the embeddings from the trained models.
We compared Spherical K-means~\cite{dhillon2001concept} and Hierarchical Agglomerative Clustering (HAC)~\cite{lance1967general}.
Cosine similarity was used as a similarity measure in both algorithms.
In Spherical K-means, the true number of clusters was given while in HAC the threshold was adjusted by the validation set.
We used Adjusted Mutual Information (AMI) ~\cite{vinh2010information}, unsupervised accuracy rate (ACC), Paired ($F_P$) and BCubed F-score ($F_B$)~\cite{amigo2009comparison} as metrics.

\subsection{Evaluation on EK-Instance Dataset}
\begin{table}[t]
\begin{center}
\scalebox{.75}{
\begin{tabular}{lrrrrr}
\hline
Model & Clustering Method & AMI & ACC & $F_P$ & $F_B$ \\ \hline
\multirow{2}{*}{ImageNet} & K-means & 0.755 & 0.590 & 0.547 & 0.628 \\ 
 & HAC & 0.849 & 0.735 & 0.724 & 0.770 \\ \hline
\multirow{2}{*}{ArcFace} & K-Means & 0.794 & 0.631 & 0.606 & 0.670 \\ 
 & HAC & 0.892 & 0.796 & 0.784 & 0.827 \\ \hline
\multirow{2}{*}{N-pair} & K-Means & 0.834 & 0.713 & 0.646 & 0.724 \\ 
 & HAC & \textbf{0.924} & \textbf{0.854} & \textbf{0.847} & \textbf{0.874} \\ \hline
\end{tabular}
}
\caption{Performance on EK-Instance Dataset.}
\label{tab:clustering_result}
\end{center}
\end{table}

Table~\ref{tab:clustering_result} shows the performance in the EK-Instance dataset.
While ImageNet without fine-tuning showed moderate performance, the fine-tuned model in the training set showed better performance.
In particular, the combination of N-pair loss and HAC showed the best performance.
The reason why HAC exceeded K-means was that the K-means method tended to generate relatively even clusters, while the actual frequency distribution was biased.

\begin{figure}[t]
\centerline{\includegraphics[width=0.75\linewidth]{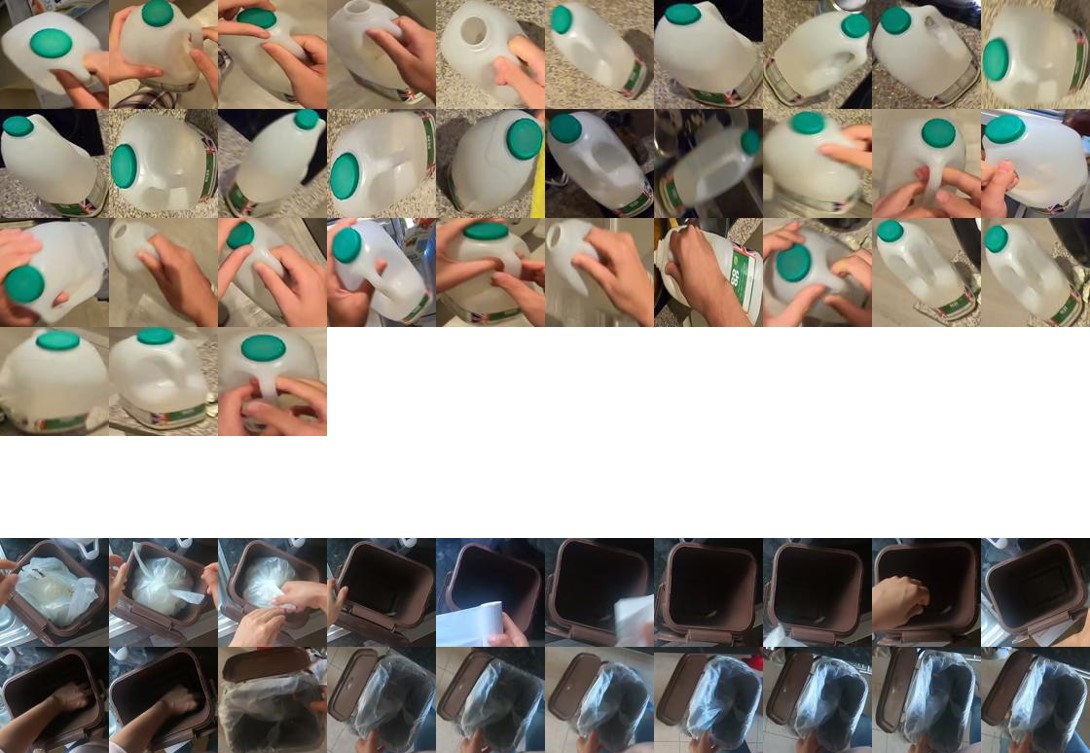}}
\vspace{-0.5em}
\caption{Example of obtained clusters.}
\label{fig:qual_1}
\end{figure}

\begin{figure}[t]
\centerline{\includegraphics[width=0.75\linewidth]{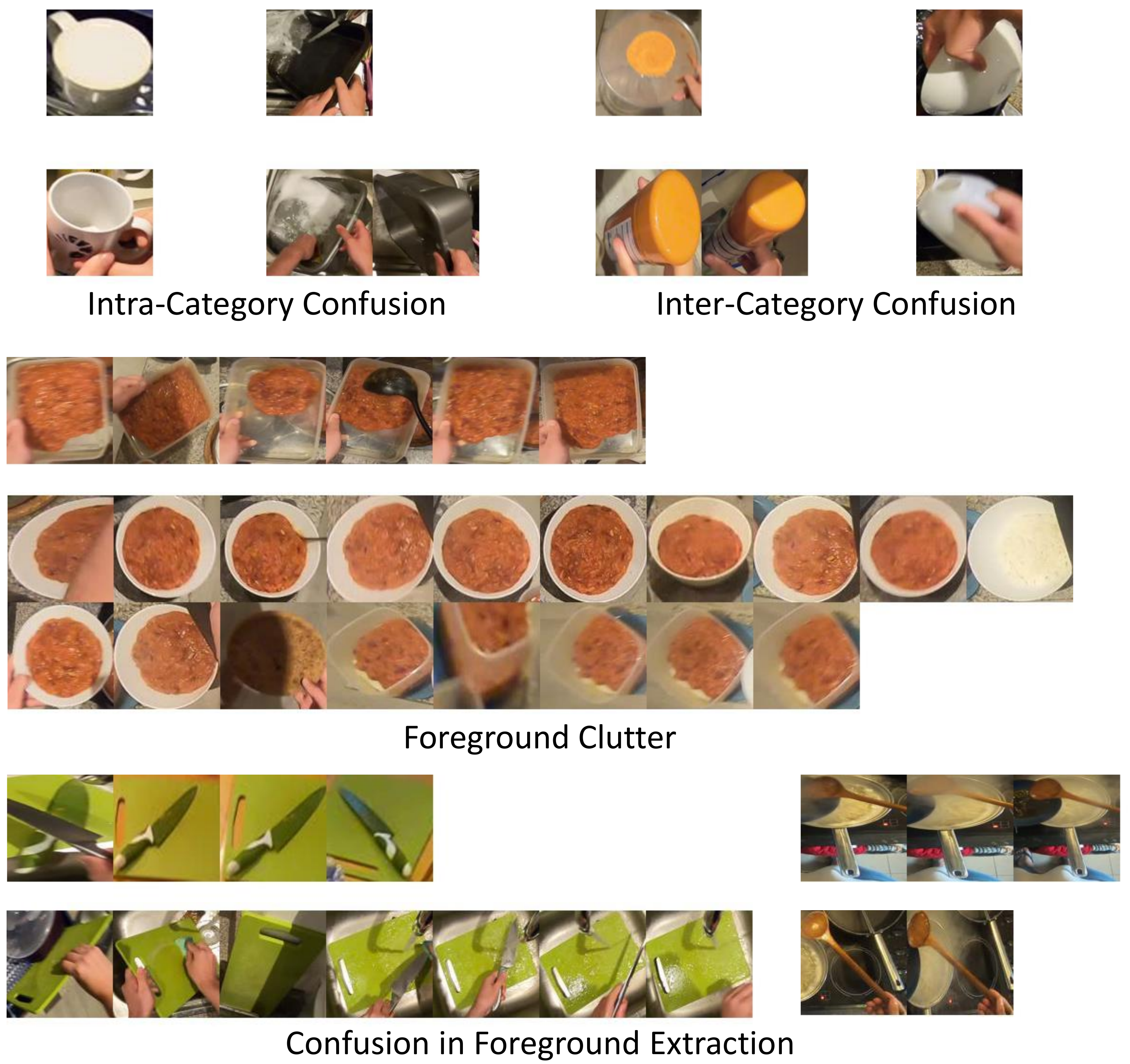}}
\vspace{-0.5em}
\caption{Failed HAC merging attepnts. First and second row of each item show tracks that are wrongly merged during HAC.}
\label{fig:hac_failure}
\end{figure}

Figure~\ref{fig:qual_1} shows an example of clusters extracted by the combination of N-pair and HAC.
The model was not only robust to changes in the pose of the object and occlusion by hand (upper) but was also able to absorb foreground clutter such as putting a plastic bag on the trash can (lower).

However, some failure cases were seen as shown in Figure~\ref{fig:hac_failure}.
First, there was intra/inter-category confusion with similar colors and textures.
In particular, there were cases where completely different types of objects were confused by superficial features.
As a unique failure in dynamic environments, confusion due to severe foreground clutter was seen when the same meal was served in another container as in the second line.
Moreover, when multiple objects overlap, the model misunderstood the target object specified by a bounding box (\eg, a cutting board and a knife on it).

\begin{table*}[t]
\setlength{\tabcolsep}{1.5mm}
\begin{center}
\scalebox{.8}{
\begin{tabular}{lrrrrrrrrrrrr}
\hline
Dataset & \multicolumn{3}{c}{EK-Instance} & \multicolumn{3}{c}{Online Products} & \multicolumn{3}{c}{INSTRE} & \multicolumn{3}{c}{CORe50} \\
 & AMI & $F_P$ & $F_B$ & AMI & $F_P$ & $F_B$ & AMI & $F_P$ & $F_B$ & AMI & $F_P$ & $F_B$ \\ \hline
ImageNet & \blue{0.849} & \blue{0.724} & \blue{0.770} & 0.332 & 0.023 & 0.434 & \blue{0.944} & 0.836 & \blue{0.890} & 0.497 & 0.311 & 0.446  \\
EK-Instance & \textbf{0.924} & \textbf{0.847} & \textbf{0.874} & \blue{0.358} & \textbf{0.360} & \blue{0.448} & 0.930 & \blue{0.837} & 0.884 & \blue{0.578} & \blue{0.415} & \blue{0.532} \\
Online Products & 0.800 & 0.620 & 0.694 & \textbf{0.649} & \blue{0.321} & \textbf{0.633} & 0.870 & 0.756 & 0.773 & 0.090 & 0.063 & 0.216  \\ 
INSTRE & 0.795 & 0.634 & 0.703 & 0.304 & 0.047 & 0.445 & \textbf{0.984} & \textbf{0.963} & \textbf{0.972} & \textbf{0.667} & \textbf{0.440} & \textbf{0.599} \\ 
\hline
\end{tabular}
}
\captionsetup{width=.8\textwidth}
\caption{Results of cross-dataset transfer. Left column shows source dataset and top row shows target dataset. Numbers in bold and blue show best and second best results across source datasets, respectively.}
\label{tab:transfer}
\end{center}
\end{table*}

\subsection{Cross-Dataset Transfer}
To investigate the properties of the learned feature representations in each dataset, we evaluated the performance when a model trained in one dataset was transferred to another dataset.
We used Stanford Online Products (SOP)~\cite{song2016deep} as a dataset using internet images, INSTRE~\cite{wang2015instre} and CORe50 ~\cite{lomonaco2017core50} as a dataset containing instances on various backgrounds. CORe50 was used only for evaluation.

As shown in Table~\ref{tab:transfer}, the models trained by SOP and INSTRE did not generalize to EK-Instance (worse than ImageNet), while the model trained by EK-Instance showed improvement in SOP and CORe50, showing that it can respond to both foreground and background clutter.

\section{Discussion}
We draw the following suggestions:
{\bf (i) Robustness against instance-specific appearance change is needed:}
The newly constructed EK-Instance dataset showed that the appearance of an object changes across interactions in a dynamic environment.
Towards practical object instance identification, it is necessary to acquire not only viewpoint invariance but also robustness to changes in appearance due to interaction.
{\bf (ii) Both low-level/high-level features are important:}
There was confusion between instances with similar low-level features such as color and texture but from completely different categories.
We suspect that the category-level information was lost during training.
A more robust approach is needed that uses both low-level and high-level features as clues for identification.
{\bf (iii) Selective foreground feature extraction is demanded:}
In this work, bounding boxes were used to indicate the target object.
However, when instances intersect with each other, confusion in determining the target object was observed.
An explicit method to selectively filter features of the target object based on contextual information will be demanded.

\section{Conclusion}
In this study, we tackled the problem of object instance identification in a dynamic environment with human-object interaction.
We proposed the EK-Instance dataset to analyze its difficulty.
Specifically, we formulated the problem as clustering of image tracks and conducted an extensive analysis by within/cross-dataset evaluation.
The analyses have revealed the existing biases in the trained model, and have discovered challenges for further development such as robustness against appearance change, feature fusion, and selective foreground feature extraction.

{\small

}

\appendix
\section{Details on EK-Instance Dataset}
\subsection{Statistics}
\begin{figure*}[t]
\centerline{\includegraphics[width=0.9\linewidth]{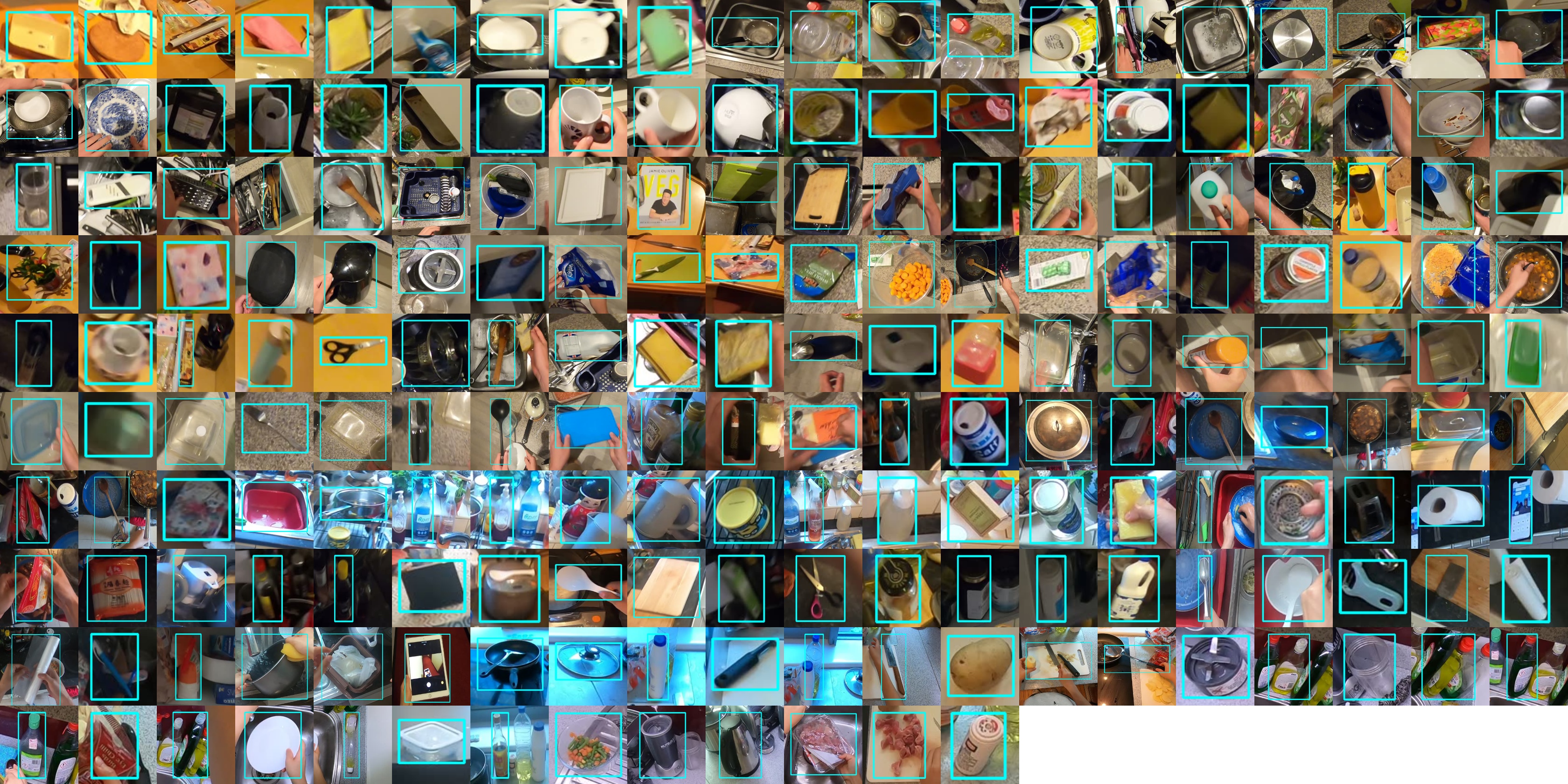}}
\caption{Images used in validation set (193 instances).}
\label{fig:valid_instances}
\end{figure*}

\begin{figure*}[t]
\centerline{\includegraphics[width=0.9\linewidth]{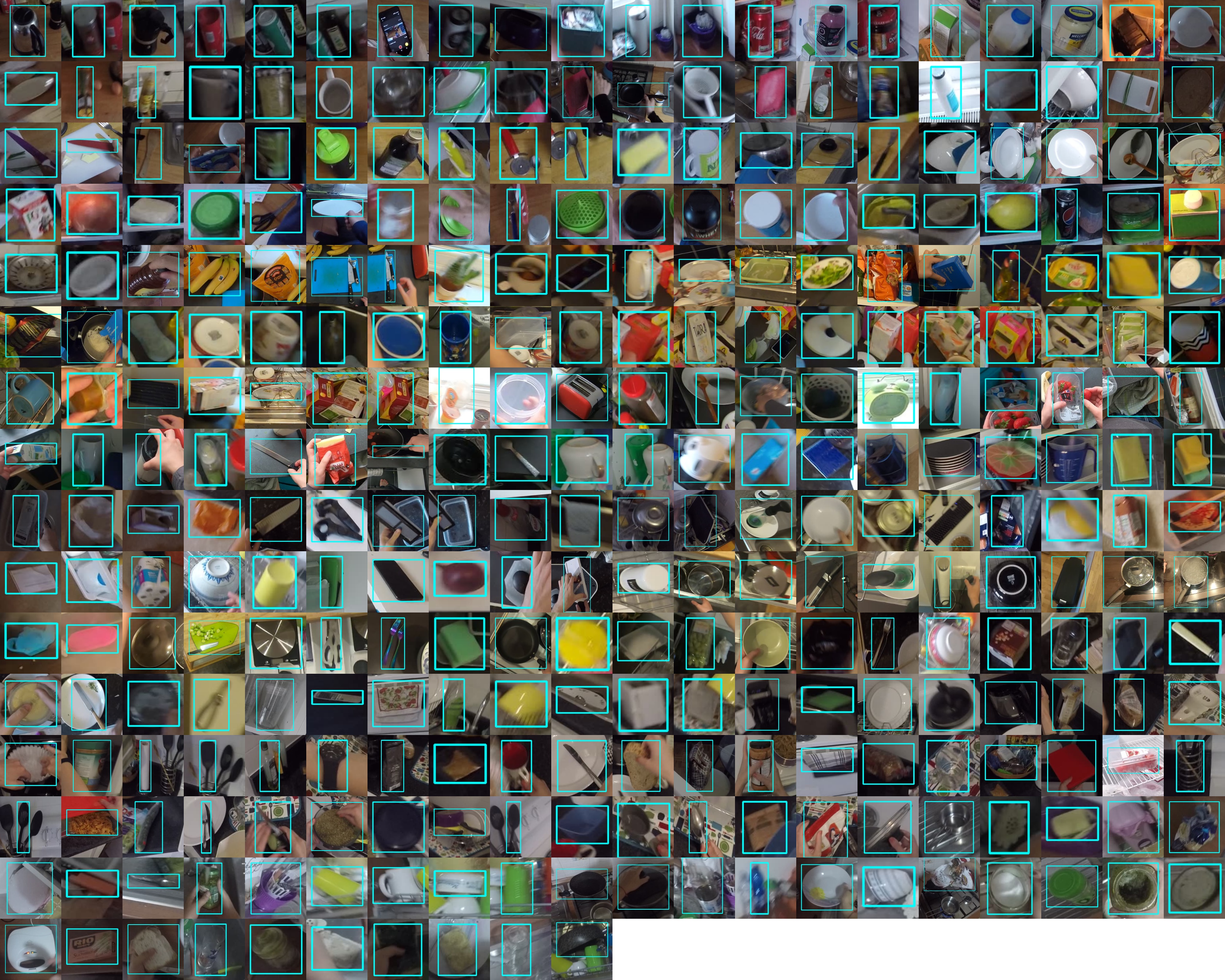}}
\caption{Images used in test set (310 instances).}
\label{fig:test_instances}
\end{figure*}

As mentioned in the main text, we collected 1,554 instances, 38,479 tracks, and 92,376 frames from 23 participants in total.
Following the split used in the EPIC-KITCHENS dataset~\cite{Damen2020RESCALING}, we split the dataset into training, validation, and test set.
Concretely, we used instances from P33, P34, and P36 for validation and P09, P11, P18, and P32 for testing.
Table~\ref{tab:ekinstance_statistics} shows the statistics of each split.
Figure~\ref{fig:valid_instances} and \ref{fig:test_instances} show the images of the instances included in validation and test set, respectively.
By collecting more than 1,500 instances, we were able to train a metric learning-based neural network model.
In addition, the subject-based split allowed us to evaluate the instance identification performance in novel environments.

\begin{figure*}[t]
\centerline{\includegraphics[width=0.6\linewidth]{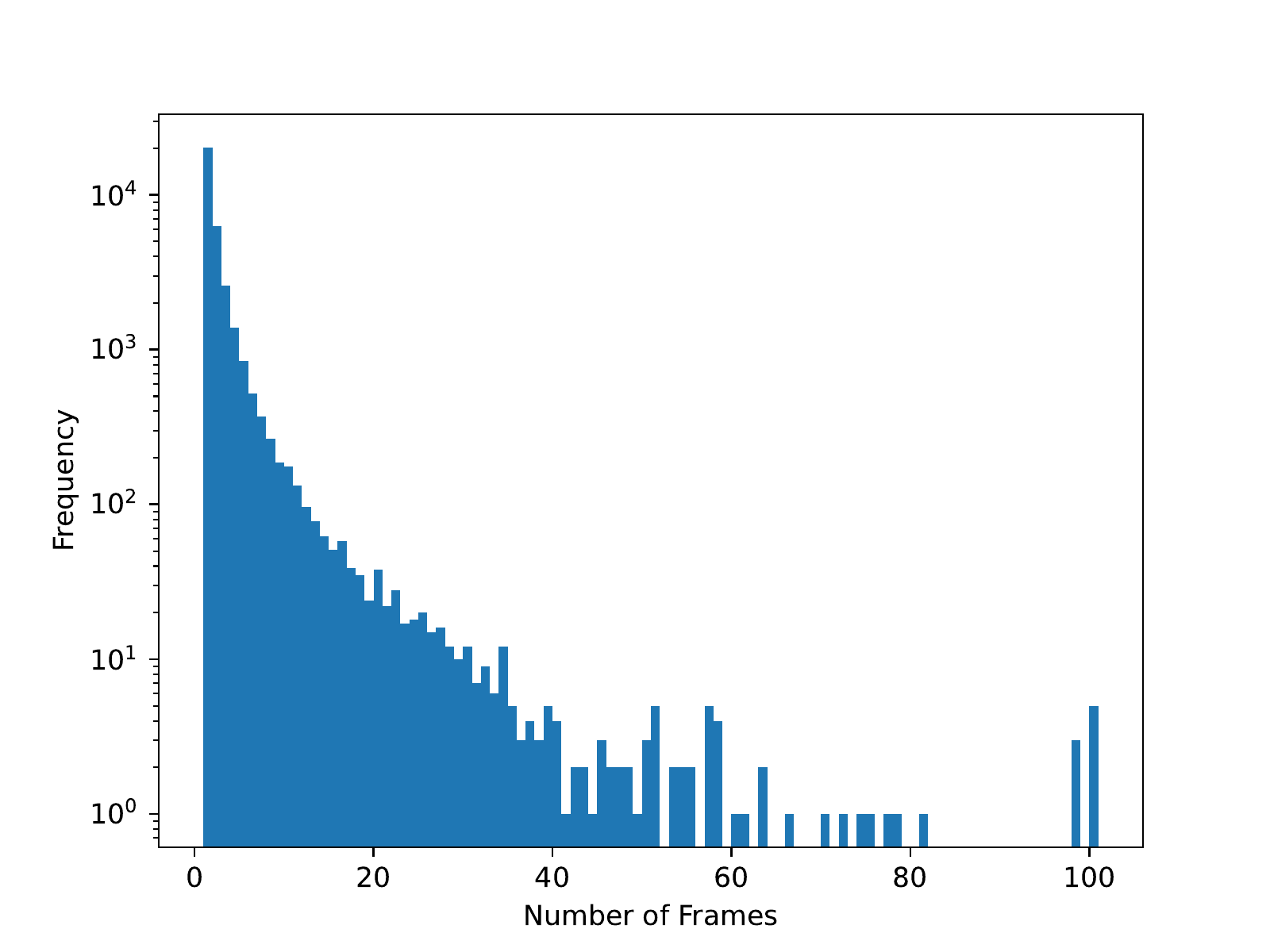}}
\caption{Distribution of number of images per track.}
\label{fig:nb_frames}
\end{figure*}

\begin{figure*}[t]
\centerline{\includegraphics[width=0.6\linewidth]{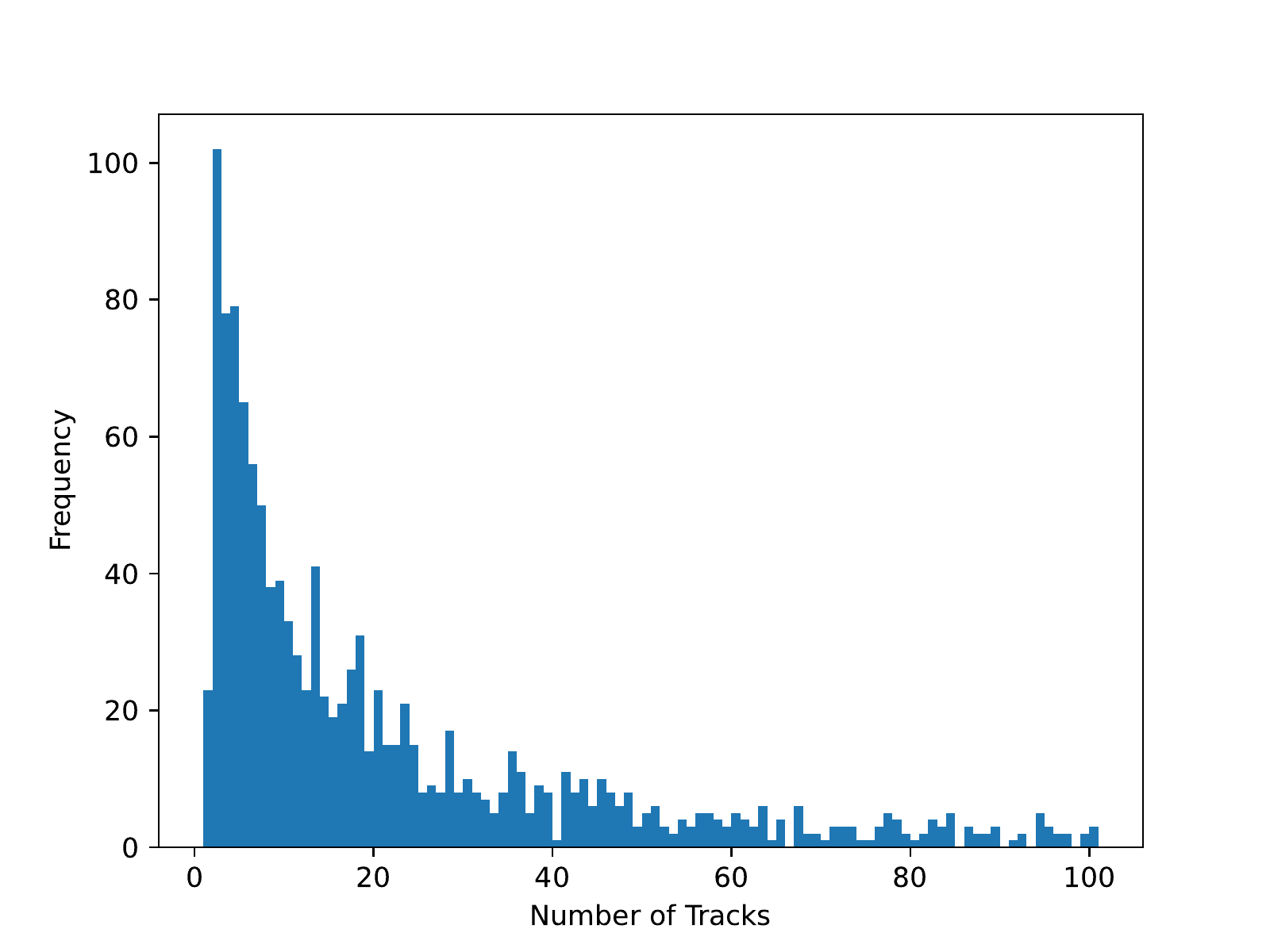}}
\caption{Distribution of number of tracks per instance.}
\label{fig:nb_tracks}
\end{figure*}

The frequency of the instances is highly biased as it reflects the occurrence rate of objects in real-world videos.
Figure~\ref{fig:nb_frames} and \ref{fig:nb_tracks} shows the frequency of the number of images per track and the number of tracks per instance, respectively.
It can be seen that most instances appear a small number of times in a very short time, whereas a very small number of instances appear a long time and a large number of times.

\begin{table*}[t]
\begin{center}
\begin{tabular}{lrrrr}
\hline
Split & \#participants & \#instances & \#tracks & \#frames  \\ \hline
Training & 16 & 1051 & 27234 & 64300 \\
Validation & 3 & 193 & 4647 & 11851 \\
Test & 4 & 310 & 6598 & 16225 \\ \hline
Total & 23 & 1554 & 38479 & 92376 \\ \hline
\end{tabular}
\caption{Statistics of each split.}
\label{tab:ekinstance_statistics}
\end{center}
\end{table*}

\subsection{Additional Examples}
\begin{figure*}[t]
\centerline{\includegraphics[width=0.65\linewidth]{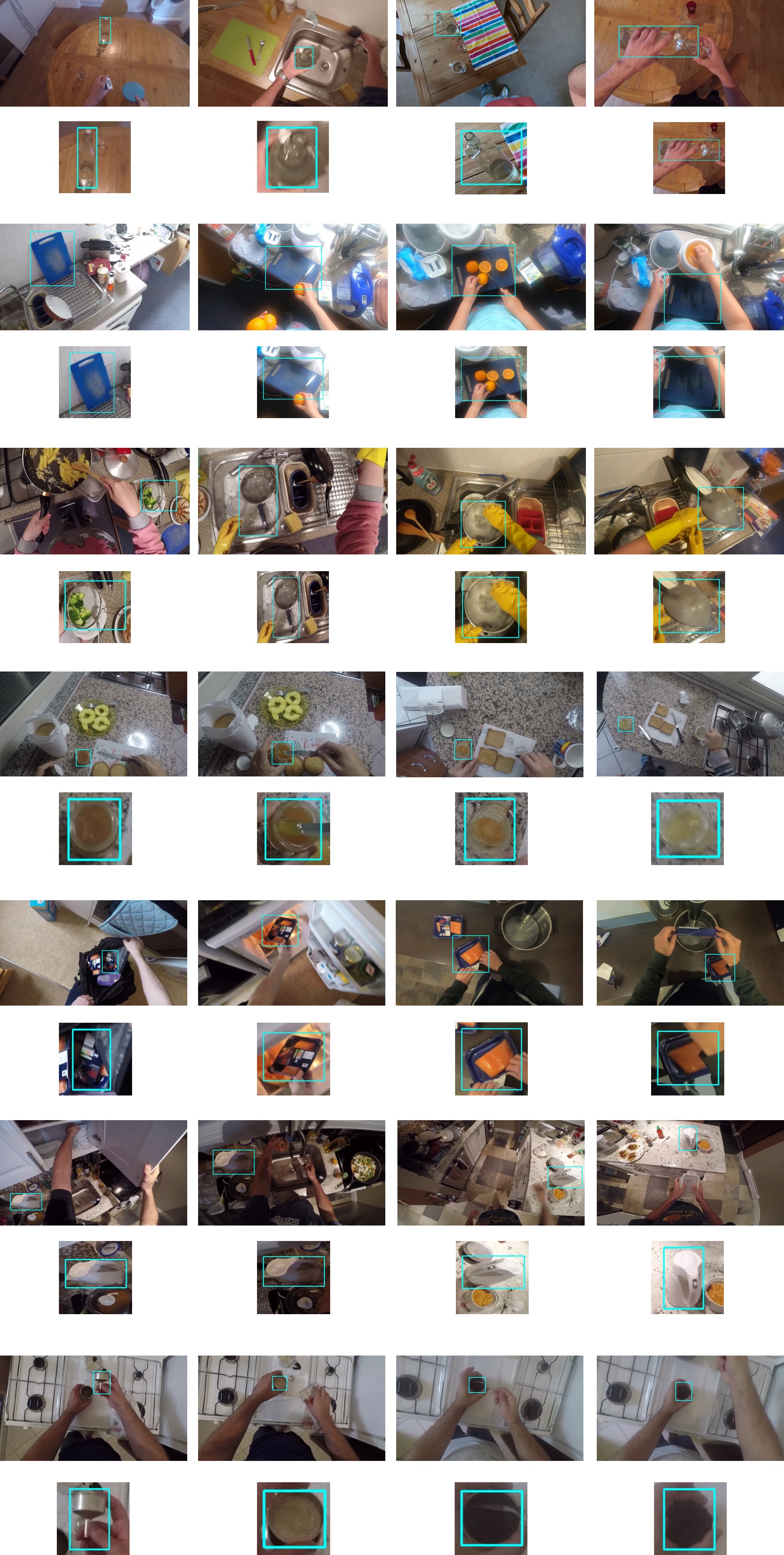}}
\caption{Additional examples of EK-Instance dataset. Each row refers to same instance.}
\label{fig:additional_examples}
\end{figure*}

Figure~\ref{fig:additional_examples} shows additional examples on the instances included in the EK-Instance dataset.
Not only changes in the posture of the object and the lighting environment, but also various changes such as disturbances caused by hands and other objects, changes in appearance due to operation (\eg, using the contents, peeling off the packaging), changes in the surrounding background, etc. are within the same instance. It is characteristic that it occurs in.

\section{Details on Baseline Methods}
\subsection{Normalized Softmax (N-Softmax) Loss~\cite{zhai2019classification}}
Normalized softmax loss is one of the simplest loss function for metric learning, extending the cross-entropy loss for classification task.
It is almost the same as the cross-entropy loss except the embedded vector and the weight vector of the final layer are L2 normalized.

$$
L_{\mathrm{nsoftmax}} = -\frac{1}{N} \sum_{i=1}^{N} \log{\frac{e^{W_{y_i}^\mathsf{T} {\bf z}'_i / \tau}}{\sum_{k=1}^K e^{W_{k}^\mathsf{T} {\bf z}'_i / \tau}}},
$$
where ${\bf z}'_i = \frac{{\bf z}_i}{\|{\bf z}_i\|}$ is a vector by applying L2-normalization to ${\bf z}_i$ and $\tau$ denotes a temperature parameter that adjusts the scale of the numerator and denominator.
We set $\tau=0.2$.

\subsection{Additive Angular Margin (ArcFace) Loss~\cite{deng2019arcface}}
ArcFace loss is described as follows:
$$
L_{\mathrm{arcface}} = -\frac{1}{N} \sum_{i=1}^{N} \log{\frac{e^{\cos{(\theta_{y_i}+m)} / \tau}}{e^{\cos{(\theta_{y_i}+m)} / \tau}+\sum_{k=1, k \neq y_i}^K e^{\cos{\theta_j} / \tau}}},
$$

where $\theta_k$ denotes the angle between L2-normalized weight $W_k$ and a feature vector ${\bf z}'_i$ which belongs to class $y_i$.
$m$ denotes a margin term to increase the inter-class distance.
This loss function tries to maintain a feature vector to have a margin of $m$ against all the feature vectors that belongs to a different class.
As a result, feature vectors of a same class will be distributed in a narrower space.
We set $m=0.5$ and $s=1/30$.

\subsection{N-pair Loss~\cite{sohn2016improved}} 
Different from the classification-based losses above, the n-pair loss tries to increase the similarity between feature vectors from the same class while decrease the similarity between feature vectors from different classes.
Especially, the n-pair loss enables efficient training by taking all the possible pairs within the batch, without using classification weight $W$.
$$
L_{\mathrm{npair}} = -\frac{1}{N} \sum_{i=1}^{N} \log{\frac{\sum_{j=1, y_j = y_i}^N e^{{\bf z}_i^\mathsf{T} {\bf z}_j / \tau}}{\sum_{j=1}^N e^{{\bf z}_i^\mathsf{T} {\bf z}_j  / \tau}}}.
$$
We set $\tau=0.07$.

\section{Details on Evaluation}
\subsection{Data Split}
We performed clustering within the images from the same participant and reported the average score among them.

\subsection{Evaluation Metrics}
We describe the four evaluation metrics in detail:

{\bf Adjusted Mutual Information (AMI)~\cite{vinh2010information}}:
A metric based on mutual information.
Given a set of ground truth clusters $C=\{c_i\}_{i=1}^N$ and predicted clusters $K=\{k_i\}_{i=1}^N$, we calculate AMI as follows:
$$
AMI = \frac{I(C, K) - \mathbb{E}[I(C, K)]}{\frac{1}{2}[H(C)+H(K)]-\mathbb{E}[I(C, K)]},
$$
$I$ and $H$ denote mutual information and entropy, respectively.
AMI will be 1.0 if the ground truth clusters and the predicted clusters are identical.

{\bf Unsupervised Accuracy (ACC)}: 
Given a set of ground truth clusters and a set of predicted clusters, the maximum accuracy can be achieved when a one-to-one correspondence between the two sets of clusters is made.
Hungarian algorithm~\cite{kuhn1955hungarian} is used to calculate the optimal assignment $m$.
$$
ACC=\max_m \frac{\sum_{i=1}^{N} \mathbf{1} \{c_i = m(k_i)\}}{N}.
$$

{\bf Paired F-score ($F_P$)}: 
Given $N$ samples, takes all the $\frac{1}{2} N (N-1)$ combinations and counts them as correct if the relationship between the two samples is the same in predicted clusters and ground truth clusters.

{\bf BCubed F-measure ($F_B$)~\cite{amigo2009comparison}}:
This metric calculates cluster-wise precision and recalls while $F_P$ takes all the pairs regardless of the predicted clusters.
Specifically, the {\it correctness} between samples $i$ and $j$ will be defined as follows:
$$
Correctness\:(i, j) = 
\begin{cases}
1 & \mathrm{if} \:L(i)\:=\:L(j)\:\mathrm{and}\:C(i)\:=\:C(j) \\
0 & \mathrm{otherwise},
\end{cases}
$$
where $L(i)$ and $C(i)$ refers to the $i$-th predicted clusters and ground truth clusters.
Finally, BCubed Precision and BCubed Recall will be calculated as follows:
$$
BCubed\:Precision = \frac{1}{N} \sum_i^N \sum_{j\in C(i)} \frac{Correctness\:(i, j)}{|C(i)|},
$$
$$
BCubed\:Recall = \frac{1}{N} \sum_i^N \sum_{j\in L(i)} \frac{Correctness\:(i, j)}{|L(i)|}.
$$

\section{Details on Datasets}
\paragraph*{Stanford Online Products~\cite{song2016deep}:}
This dataset is composed of 120,053 images from 22,634 online product categories collected from eBay.
Following the original split, we used 11,318 categories for training and 11,316 categories for testing, while the hyperparameter was tuned by further partitioning the training set for validation.

\paragraph*{INSTRE~\cite{wang2015instre}:}
This dataset consists of distinctive objects such as architecture, planar objects, and toys.
The Internet images and manually collected images are mixed and divided into three subsets based on filming conditions.
Notably, manually collected images were intentionally shot in 25 different backgrounds.
For the evaluation, we used 100 instances from the INSTRE-S1 subset which contains 11,011 images, and 50 instances each are used for training and testing.

\paragraph*{CORe50~\cite{lomonaco2017core50}:}
This dataset contains 50 hand-held objects belonging to 10 categories such as scissors.
Although the number of instances is as small as 50, we used as a target data set because it was intentionally photographed in various backgrounds with hand occlusion, which shares the difficulties in our EK-Instance dataset.
Since this data set consists of videos for a total of 550 trials, images were extracted at 1~fps to compose a image track.
Since the bounding box of the object attached to the data set was cut out looser than the target object, we crop 28 pixels at each of the four corners of the original 128-pixel square bounding box to form an input.

\section{Visualization of Learned Features}
\begin{figure*}[t]
\centerline{\includegraphics[width=0.8\linewidth]{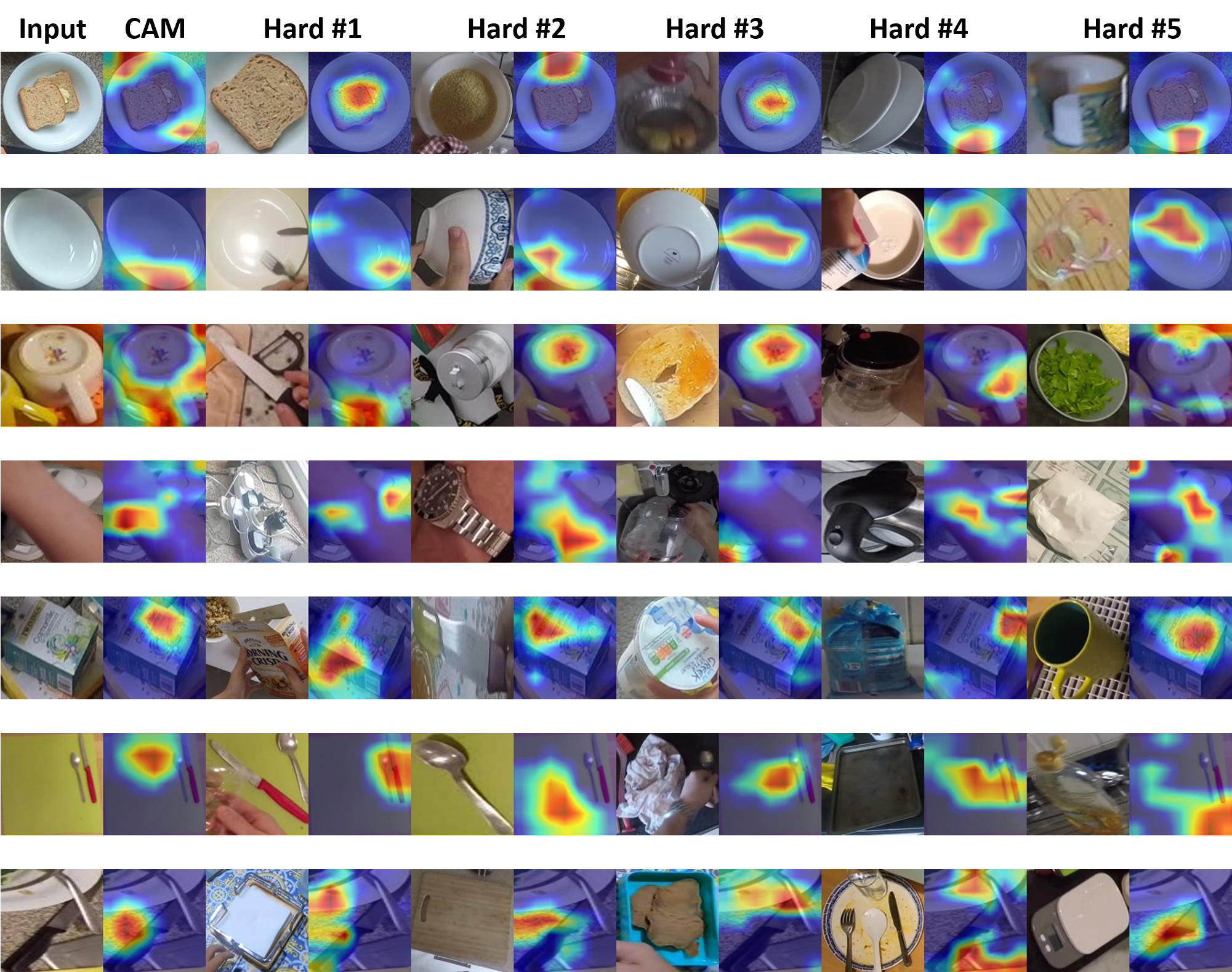}}
\caption{Grad-CAM visualization on training set. First and second column are input image and its support for ground-truth category. Other columns are top-5 ``hard'' classes which showed high instance classification probability and their corresponding supports, respectively.}
\label{fig:grad_cam}
\end{figure*}

To understand which region the model is focusing on, we applied Grad-CAM~\cite{selvaraju2017grad} and visualized the activation heatmap of each predicted instance (class).
For simplicity, we used the model trained with N-softmax loss.
Figure~\ref{fig:grad_cam} shows the results of discriminative regions taken from the last layer of the ResNet-34 backbone.

First, we found that the model focused on the peripheral region of specific objects such as dishes and mugs (1-3rd row).
Such unchanged regions may be discovered during the training since such objects typically have foreground clutters on their centers.
Similarly, the model focused on regions other than hands that are not discriminative for instance identification (4th row).
On the other hand, unreasonably sharp regions are generated in some instances even though the entire region contains unique patterns (5th and 6th row).
Furthermore, as mentioned in the failure examples, the model focused on backgrounds in some instances, leaving room for improvement.

\end{document}